\documentclass[11pt,a4paper]{article}
\usepackage[hyperref]{acl2021}
\usepackage{times}
\usepackage{latexsym}
\usepackage{tabularx}
\usepackage{graphicx}

\usepackage{microtype}

\aclfinalcopy % Uncomment this line for the final submission
\title{NLP-IIS@UT at SemEval-2021 Task 4: Machine Reading Comprehension using the Long Document Transformer}

\author{Hossein Basafa \\
  College of ECE \\
  University of Tehran \\
  \texttt{hbasafa@ut.ac.ir} \\
  \And
  Sajad Movahedi \\
  College of ECE \\
  University of Tehran \\
  \texttt{s.movahedi@ut.ac.ir} \\
  \AND
  Ali Ebrahimi \\
  College of ECE \\
  University of Tehran \\
  \texttt{ali96ebrahimi@ut.ac.ir} \\
  \And
  Azadeh Shakery \\
  College of ECE \\
  University of Tehran \\
  \texttt{shakery@ut.ac.ir} \\
  \And
  Heshaam Faili \\
  College of ECE \\
  University of Tehran \\
  \texttt{hfaili@ut.ac.ir} \\
  }

\date{}

\begin{document}
\maketitle
\begin{abstract}
% This paper presents a technical report of our submission to the 4th task of SemEval-2021, titled: Reading Comprehension of Abstract Meaning. We participated in subtask 2 of the task during the evaluation phase, although here we present the result of both subtasks. We reformulated the problem as the problem introduced in the WikiHop dataset \cite{welbl-etal-2018-constructing}. In WikiHop, the model is presented with multiple documents (called evidence), each of which may contain valuable information about the question. Then the model must choose the correct answer from multiple choices. Furthermore, we used the longformer model in order to better process the documents, which usually are very lengthy and require a large receptive field from the model. We improved the performance on the data from (23.01\% and 22.95\%) achieved by the baselines for subtask 1 and 2, respectively, to (70.30\% and 64.38\%).

% field and subject
This paper presents a technical report of our submission to the 4th task of SemEval-2021, titled: Reading Comprehension of Abstract Meaning.
\mbox

% state of the problem
In this task, we want to predict the correct answer based on a question given a context. Usually, contexts are very lengthy and require a large receptive field from the model. Thus, common contextualized language models like BERT miss fine representation and performance due to the limited capacity of the input tokens.
\mbox

% innovations
To tackle this problem, we used the longformer model to better process the sequences. Furthermore, we utilized the method proposed in the longformer benchmark on wikihop dataset which
\mbox

% evaluations
 improved the accuracy on our task data from (23.01\% and 22.95\%) achieved by the baselines for subtask 1 and 2, respectively, to (70.30\% and 64.38\%).

\end{abstract}

\section{Introduction}

% introduction and the field
Reading comprehension is the ability to understand a passage either by human or machine. One of the great benchmarks to evaluate this ability is to try to answer specific questions related to the passage \cite{DBLP:conf/emnlp/RajpurkarZLL16}. Generally, this problem can contain single or multiple documents as context (containing relevant information needed to understand and answer the question), a question (a sentence with at least one asking parameter), and an answer (which is the parameter value of the question).

% task description
In the Task of Reading Comprehension of Abstract Meaning (ReCAM), we have one passage as a context, one question and five candidate answers \cite{zheng-2021-semeval-task4}. The goal is to identify the correct answer based on the context and the given question. You can see a sample of the data in Table \ref{table_data}. For each instance of the data, there is a passage, a question with a missing word that should be filled based on the passage, and five candidate answers to the question.

\begin{table}[]
    \centering
    \begin{tabularx}{0.5\textwidth}{|l|X|}
    \hline
        \textbf{Passage} & ... observers have even named it after him, ``Abenomics''. It is based on three key pillars - the ``three arrows'' of monetary policy, fiscal stimulus and structural reforms in order to ensure long-term sustainable growth in the world's third-largest economy. In this weekend's upper house elections ... \\\hline
        \textbf{Question} & Abenomics: The \textit{\textbf{@Placeholder}} and the risks\\\hline
        \textbf{Answer} & (A) chances ~ (B) prospective ~ (C) security ~ \textbf{(D) objectives} ~ (E) threats\\\hline
    \end{tabularx}
    \caption{\label{table_data} An instance of the data.}
\end{table}

% subtasks descriptions
The task divides into two subtasks: imperceptibility and non-specificity\cite{zheng-2021-semeval-task4}.
\begin{itemize}
    \item imperceptibility: this level of abstract words refers to ideas and concepts that are distant from immediate perception; such as culture, economics, and politics.
    \item non-specificity: In contrast to concrete words, this subtask includes more abstract words which focus on a different type of definition; for example, a concrete word like `cow` could be interpreted as an `animal` which is considered as a more abstract word \cite{changizi2008economically}.
\end{itemize}

% problems and limitations
The main challenges of this task are the abstract meaning concept representation as well as the machine reading comprehension. This is the main reason we have utilized contextualized language representation models to tackle abstract meaning representation problems.

% solutions and innovations
In this paper, we use an end-to-end deep contextualized architecture to model this task. This model is also capable of considering more than one passage as the context, and more than five candidate answers. Since we use the long document transformer model (Longformer \cite{beltagy2020longformer}), no limitation is considered in context passage length. We have evaluated this model both on subtask-1 and subtask-2 which resulted in 70\% and 64\% accuracy, respectively. Therefore, we have about 40\% improvement compared to the baseline, which is a Gated Attention (GA) model \cite{zheng-2021-semeval-task4}.

% This is a very long sequence of words and a simple transformer-based model like BERT cannot be used to encode such sequence. To have an end-to-end unified model, we utilized Longformer which is capable of encoding long documents.

% outline
The rest of the paper is as follows: Section 2 describes the related works and the background. Section 3 includes the description of the proposed method. Section 4 contains the evaluation metrics used as well as a brief discussion, which is then followed by a conclusion and future works in section 5.

\section{Background and Related Works}
Many approaches have been presented in the literature, from pipeline-based models to end-to-end ones. Each module is also well-investigated from rule-based models to deep learning ones. Despite various configurations presented in the literature to model this problem, most of the systems consist of three modules\cite{DBLP:journals/corr/abs-2001-01582}: 
\begin{itemize}
    \item Language representation: this module is responsible to encode the inputs. Context, question, and answer need to be represented as numeric values for computational algorithms to be usable on them. Dense vectorized representations are the most popular methods, which allow us to use the majority of machine learning algorithms. 
    \item Reasoning: this module is used to find demonstrations of why the answer is assumed to be valid. It can also be used as a limiter for searchable context.
    \item Prediction: this module aims to generate, retrieve or select the correct answer based on the task description.
\end{itemize}

Recent studies are provided as follows with respect to these modules that the last two modules have been merged. In the end, the longformer model is presented as our mainstay in this paper.

\subsection{Word and text representation}
\par One of the most important problems in NLP is representation learning. 
The earliest models for word representation in the time of deep learning were the models
proposed in \cite{pennington2014glove} and \cite{mikolov2013distributed}, which utilized the weights
learned for an auxiliary task (a simplified version of the task of language modeling) for word representation. Similarly, methods proposed in \cite{le2014distributed} and \cite{liu2015topical} utilized
a similar structure for sentence, paragraph, or document representation learning.
\par While these methods were quite effective, it has been shown that using neural language models
as a way of word representation results in much better, and context-aware representations. In
\cite{howard2018universal} it has been shown that fine-tuning language models as sentence encoders
result in a significant performance improvement. At the same time, \cite{peters2018deep} used language
models directly as word representations, which resulted in significant improvements. In \cite{devlin2018bert}
a transformer model was trained for the task of masked language models, which resulted in significant improvements,
surpassing human performance in many NLP tasks. One of the shortcomings of transformers is the lack of 
a memory mechanism, which results in (theoretically) lower receptive field compared with LSTMs \cite{beltagy2020longformer}
this shortcoming was addressed by improving the self attention mechanism in transformers so that it would
have a (theoretically) unbounded receptive field. More details are presented later in this section.

\subsection{Natural language understanding}
\par Natural language understanding (NLU) is an umbrella term, referring to any tasks that require machine comprehension.
Compared to other NLP tasks, NLU requires the model to be able to understand and reason about the data \cite{semaan2012natural}.
While great progress has been made in this field by using contextual word representation \cite{devlin2018bert},
it has been found that designing the model itself must not be neglected \cite{zhu2018sdnet}. On the other hand, 
it has been shown that utilizing a transfer learning setting to share knowledge 
between different NLU tasks results in better performance with fewer data and fewer parameters \cite{pilault2020conditionally},
which proves a significant similarity between these tasks.

\subsection{The Longformer}
% longformer desc
Deep contextualized language models like BERT\cite{devlin-etal-2019-bert} have been well investigated in the literature and achieved state-of-the-art results on various tasks. However, these models suffer from performance limitations due to their self-attention layer which results in quadratic space and time complexity concerning the sequence length. In contrast, this model removes the self-attention layer from the base language models, so the limitation resolves and the complexity scales to linear. In order to increase the quality of the model compared to basic models, they have added a global attention layer to the model end which significantly outperforms state-of-the-art models on long document (passage) tasks and competitive on normal documents. Also, this configuration increases the performance on both normal and lengthy inputs which makes it a good alternative for tasks containing large inputs. This model is also evaluated on a similar task on WikiHop dataset\cite{welbl2018constructing} and improved the results in terms of accuracy\cite{beltagy2020longformer}.

\section{Method}
As mentioned in section 1, given a passage, a question, and a set of answers to the question, the goal is to predict the correct answer among the candidates, which can be seen as a benchmark to evaluate how well the model can comprehend the abstract meaning. To do so, we considered an end-to-end deep learning architecture based on the transformer architecture.

\begin{figure*}
    \centering
    \includegraphics[width=0.46\textwidth]{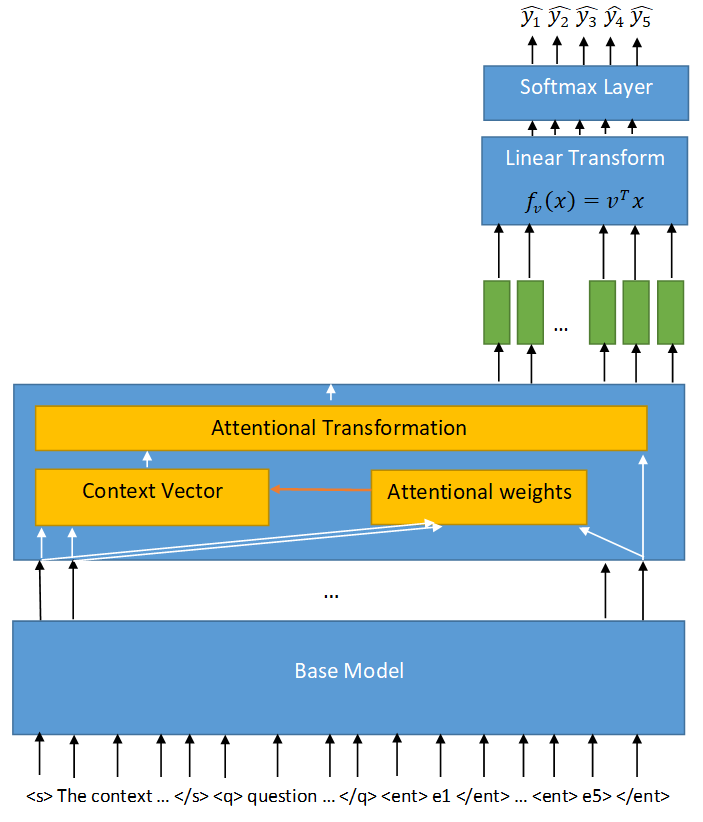}
    \caption{The model architecture. The concatenated input vector will be encoded using the base model (like RoBERTa without the self-attention layer, in our case). A global attention\cite{luong2015effective} will be applied to the question and the candidate answers representations with respect to the passage as the context. The logit (score) of each ent token will be calculated using a linear transformation function, then the prediction distribution over the answer candidates (ent tokens) will be outputted using a softmax layer.}
    \label{fig:recam_model}
\end{figure*}

Specifically, we used contextual word embeddings based on the transformer to better discover and encode the information contained in the passage. In our model, both subtasks use the same architecture as shown in figure \ref{fig:recam_model}, although we did not experiment on the possibility of multi-task learning. The word representation models are fine-tuned on the data for better performance. The fine-tuning procedure could allow us to extract additional, task-related information which could result in better accuracy in the evaluation phase.

% To find the correct answer among the candidates, we propose to modify the question by replacing the placeholder token with the candidates. This way, we will have five sentences, one of which is semantically related to the passage.

To model this problem, let $c=\{c_1, c_2, ..., c_I\}$ denote the passage as the context, where $c_i$ corresponds to the $i^{th}$ token (word or subword, depending on the tokenization technique used) and $I$ is the number of tokens in the passage. Similarly, the question is considered as $q=\{q_1, q_2, ..., q_{K}\}$ where $K$ denotes the length of the question, and $q_k$ corresponds to the $k^{th}$ token of the question. Each answer also denotes as $e^{j}$ which is only one abstract word ($j \in \{1, 2, ..., 5\}$). Then we concatenate the question and the candidates as:
\begin{equation}
	a = [q; e^{1}; e^{2}; ... ; e^{5}].
\end{equation}
The size of this sequence is $A = K + 5$ as we have only 5 candidates. Generally, this can be an arbitrary length based on the dataset.

% Specifically, let $c=\{c_1, c_2, ..., c_n\}$ denote the passage, where $c_i$ corresponds to the $i^{th}$ token (word or subword, depending on the tokenization technique used) and $n$ is the number of tokens in the passage, and $q^{j}=\{q^{j}_1, q^{j}_2, ..., q^{j}_{m}\}$ denote the question with the placeholder token replace by the $j^{th}$ answer, where $m$ denotes the length of the question. Then we concatenate the context and the candidates as:
% \begin{equation}
% 	a = [c; q^{1}; q^{2}; ... ; q^{5}].
% \end{equation}

Note that we introduce special tokens to separate the context, the question, and the candidates, similar to \cite{beltagy2020longformer}. Specifically, we introduce the tokens \verb|<s>| and \verb|</s>| for separating the context, \verb|<q>| and \verb|</q>| for separating the the question, and the tokens \verb|<ent>| and \verb|</ent>| for separating the candidates from each other. In the case of multiple passages, all passages are concatenated to form a single context. These tokens are randomly initialized and fine-tuned.

We used the Longformer model introduced in \cite{beltagy2020longformer} as the pre-trained contextual embedding model in our method. Since the context could be too long, we split the context sequence to separate chunks. Each chunk length is equal to maximum sequence length the model could accept appending the sequence $a$; in fact, $model\_max\_length= len(chunk) + len(a)$. If $c^l$ denote each chunk, this sequence could be showed as:
\begin{equation}
	b = [c^l; a]
\end{equation}
where the full context is $c=\{c^1, c^2, ..., c^L\}$, and $L$ is the last chunk. The size of this sequence is $B$ so $B = L + A$.

After feeding the input $b$ to the Longformer model, we apply a global attention only on $a$ (concatenated question and answer candidates), and the rest is the context. As the longformer model utilizes a base model (like RoBERTa without the self-attention layer, in our case), we denote this as $basemodel$ function that outputs the encoded sequence of the input. If $GAttn$ denotes the global attention function, we have:
\begin{equation}
	d_i = basemodel(b)
\end{equation}
\begin{equation}
	g_i = GAttn(d_i).1(i \in A)
\end{equation}
where $d_i$ is the raw output vector for each input token. The global attention function is applied if it is a question or answer candidate token. Then, we extract the outputs corresponding to the question and the candidates tokens, i.e. we have:
\begin{equation}
		h_j = GAttn(a,c^l)
\end{equation}

Finally, we obtain the logit of each candidate (\verb|<ent>| tokens) as $x_j$ ($x_j = h_j$ if $j$ correspond to a candidate), average over different chunks, and apply a linear transformation:
\begin{equation}
		f_j = v^T x_j
\end{equation}
where the vector $v$ is trainable, and $f_j$ is the score of each candidate. And the probability distribution over the candidates will be calculated using a softmax layer on the logits. The predicted answer is the argmax of the softmax output. we fine-tuned the model using the cross-entropy loss.

% \begin{equation}
% 		ent_j = b[\textbf{idx}_j]
% \end{equation}
% where $\textbf{idx}_j$ corresponds to the index of the \verb|<ent>| token of the $j^{th}$ candidate. Finally, we used a trainable weight $u$ to find the logit corresponding to each candidate: $l_j = u^T ent_j$. We apply a softmax layer on the logits $l_j$ to find the likelihood of each candidate being the answer. Finally, we fine-tuned the model using the cross-entropy loss.

%, where $f(.; .)$ corresponds to the Longformer model along with the attention layer and $\theta$, corresponds to their trainable weights.

\section{Evaluation}

\begin{table*}
\centering
\begin{tabular}{lrrr}
\hline \textbf{Metrics} & \textbf{Baseline(GA)} & \textbf{BERT} & \textbf{Our Method} \\ \hline
% Train Loss & - & 0.036 \\
% Val Loss & 1.603 & 1.43 \\
% Val Accuracy & 24.01\% & 72.20\% \\
% Val Macro Avg F1 & 0.234 & - \\
% Val Weighted Avg F1 & 0.234 & - \\
% Loss & 1.606 & - \\
Accuracy & 23.01\% & 63.43\% & \textbf{70.30\%} \\
Macro Avg F1 & 22.83\% & 63.38\% & \textbf{70.23\%} \\
Weighted Avg F1 & 22.76\% & 63.40\% & \textbf{70.27\%} \\
\hline
\end{tabular}
\caption{\label{table_subtask1} Subtask1 evaluation metrics on the test set }
\end{table*}

\begin{table*}
\centering
\begin{tabular}{lrrr}
\hline \textbf{Metrics} & \textbf{Baseline(GA)} & \textbf{BERT} & \textbf{Our Method} \\ \hline
% Train Loss & - & 0.051 \\
% Val Loss & 1.608 & 1.67 \\
% Val Accuracy & 21.62\% & 65.6\% \\
% Val Macro Avg F1 & 0.206 & - \\
% Val Weighted Avg F1 & 0.207 & - \\
% Loss & 1.601 & - \\
Accuracy & 22.95\% & 58.76\% & \textbf{64.38\%} \\
Macro Avg F1 & 22.42\% & 58.72\% & \textbf{64.35\%} \\
Weighted Avg F1 & 22.45\% & 58.75\% & \textbf{64.40\%} \\
\hline
\end{tabular}
\caption{\label{table_subtask2} Subtask-2 evaluation metrics on the test set}
\end{table*}

Although we only participated in the second subtask, we will evaluate our model on both subtasks here. We will explain our configurations for utilizing the model on the task as well as other baselines which are the BERT-base as an alternative model and the Gate-Attention (GA) as our task baseline. Finally, a brief discussion will be done based on the results.

\subsection{Metrics}
% metrics
Popular metrics to evaluate these models are F1, EM (Exact Match or accuracy), and MRR (Mean Reciprocal Rank). As the precision and recall in our task are equal, so F1 = Precision = Recall. Also, F1 and EM are the same. And, the use of MRR is optional, so the metrics used to evaluate the result are the accuracy and the F1.

\subsection{Baseline configuration}
% baseline configs
\par The baseline model (GA) is trained for 30 epochs, each epoch containing 101 mini-batches. The train batch size is set to 32. Dropout with the rate of 0.5 is also applied to the hidden states, and the learning rate is set to 0.001. The dimensionality of the GloVe embedding is 300, and the hidden size is set to 128. Training and evaluation take about 2 hours on a single v100 GPU. 

\subsection{BERT configuration}
% competetive configs
We use the same configuration as our method except for the global attention mechanism. In fact, we consider the output vector of each chunk as our final vector to be linearly transformed into single logit, followed by a softmax layer using the cross-entropy loss. Similarly, the logit is averaged over different chunks, before applying the linear transformation. Note that the maximum sequence length here is bounded to 512 tokens, and the model includes the $n^2$ attention mechanism. We use the base version of the model and fine-tuned it on each subtask.

\subsection{Our method configuration}
% method configs
\par We used the same model introduced in section 3 for both subtasks. The model was initialized using the Longformer-base pre-training weights, then fine-tuned in each of the subtasks. Due to the performance issues, the model max sequence length is set to 4096 tokens which are sufficient in our case. We also used the RoBERTa-large tokenizer to tokenize the input sequence as the Longformer model has been trained on using this configuration. We used a batch size of 32 and a maximum learning rate of 3e-5 using the Adam optimizer with beta2=0.98. We then assumed the validation check interval to 250 which indicates the number of gradient updates between checking validation loss. And a weight decay of 0.01 has been considered to regularize the model and avoid overfitting.

\par Our proposed model is trained for 15 epochs for each task. Fine-tuning the model takes about six hours, and inference takes about nine seconds for each sample on a single V100 GPU.

\subsection{Evaluation od Subtask 1}
Subtask1 measures imperceptibility abstract level of language understanding. This subtask includes 3227 training samples, 837 validation samples, and 2025 test samples. The size of the biggest sample in terms of context length is about 2000 tokens. We have achieved an accuracy of 70\% on the validation set, which improves our baseline by about 40 percent. Table \ref{table_subtask1} showed the results of this subtask.

\subsection{Evaluation on Subtask 2}
Subtask2 measures the non-specificity level of abstract meaning in reading comprehension. It includes 3318 training samples, 851 validation samples, and 2017 test samples. The best accuracy on the validation set is 64\%.  Table \ref{table_subtask2} showed the results of this subtask.

\subsection{Discussion}
We used two baselines to find out the effect of using a pre-trained model rather than a simple RNN model. Although this task offers a higher level of representation, using the pre-train models is helpful, and there is a higher chance of modeling such abstract concepts.

The results on subtask2 are weaker than subtask1 in pre-trained models. This can be the consequence of limited semantic representation for abstract word which indicates the subtask2 includes more abstract words in terms of abstract level; for example, the word 'animal' could be matched to any animal, like 'cat' or 'dog', but the word 'entity' is hard to be represented as it could be matched to a large number of words. And the model faces a limitation in the knowledge representation. Another assumption could be the data enrichment that these model has been trained on. As most of the available texts for training consist of concrete words, it is more likely to leverage the language understanding to less abstract words to achieve a better result.

Comparing our method which is based on longformer model to usual language models like BERT indicates a new insight in terms of passage length and the attention mechanism. Popular language models like BERT and RoBERTa use a $n^2$ attention which requires a large receptive field to represent long passages. This results in the performance limitation which bounds the input sequence up to 512 tokens. In contrast, the longformer global attention mechanism relaxes this limitation as we only need to pay attention to a small factor of context and more focus on the local window. So the receptive field will not overflow and saves the necessary information to better represent the language.

We have analyzed the errors that mostly affect our model performance. We think that the problem is the contextual representation of the language modeling, which is not well-suited in our method i.e. concatenating the context, question, and answer. The main disadvantage of concatenating the candidate answers to each other is the missing fine contextual representation as the state-of-the-art models consume the position embedding. Additionally, incorrect candidates register additional noise to each word representation as well as the placeholder in the question. 

\section{Conclusion and Future works}
We have shown how different approaches can be leveraged to machine reading comprehension of abstract meaning. We reformulated the longformer model to learn abstract meaning as a new level of semantic in machine reading comprehension. This method can also be improved by taking advantage of external knowledge and task-specific model architectures that optimize the current baseline. 

% \section*{Acknowledgments}

% The acknowledgments should go immediately before the references. Do not number the acknowledgments section.
% \textbf{Do not include this section when submitting your paper for review.}

\bibliographystyle{acl_natbib}
\bibliography{anthology,acl2021}

%\appendix

\end{document}